\documentclass[conference]{IEEEtran}
\usepackage{blindtext, graphicx}
\ifCLASSINFOpdf
\else
\fi
\usepackage[cmex10]{amsmath}
\usepackage{amssymb}

\usepackage[font=footnotesize,caption=false]{subfig}

\usepackage{todonotes}
\usepackage{gensymb}
\usepackage{bm}

\begin{document}
%
\title{Fitting 3D Morphable Models using Local Features}

\author{\IEEEauthorblockN{Patrik Huber, Zhen-Hua Feng, William Christmas, Josef Kittler}
\IEEEauthorblockA{Centre for Vision, Speech and Signal Processing\\
University of Surrey, 
Guildford GU2 7XH, UK\\
Email: \{p.huber, z.feng, w.christmas, j.kittler\}@surrey.ac.uk}
\and
\IEEEauthorblockN{Matthias R\"atsch}
\IEEEauthorblockA{Reutlingen University\\
D-72762 Reutlingen, Germany\\
Email: matthias.raetsch@reutlingen-university.de}
}


%

\maketitle

\begin{abstract}
In this paper, we propose a novel fitting method that uses local image features to fit a 3D Morphable Model to 2D images.
To overcome the obstacle of optimising a cost function that contains a non-differentiable feature extraction operator, we use a learning-based cascaded regression method that learns the gradient direction from data. The method allows to simultaneously solve for shape and pose parameters. Our method is thoroughly evaluated on Morphable Model generated data and first results on real data are presented. Compared to traditional fitting methods, which use simple raw features like pixel colour or edge maps, local features have been shown to be much more robust against variations in imaging conditions.
Our approach is unique in that we are the first to use local features to fit a Morphable Model.

\vspace{-1mm}
Because of the speed of our method, it is applicable for real-time applications. Our cascaded regression framework is available as an open source library\footnote{https://github.com/patrikhuber}.

\end{abstract}


%
\IEEEpeerreviewmaketitle

\section{Introduction}

This work tackles the problem of obtaining a 3D representation of a face from a single 2D image, which is an inherently ill-posed problem with many free parameters: the orientation of the face, identity and lighting, amongst others. A 3D Morphable Face Model (3DMM)~\cite{blanz1999morphable,blanz2003face} usually consists of a PCA model of shape and one of color (albedo), camera and a lighting model. To fit the model to a 2D image, or in other words, to reconstruct all these model parameters, a cost function is set up and optimised.

Existing so-called fitting algorithms that define and solve these cost functions can loosely be classified into two categories: the ones with linear cost functions and those with non-linear cost functions. Algorithms that fall into the first category typically use only facial landmarks like eye or mouth corners to fit shape and camera parameters, and use image information (pixel values) to fit an albedo and light model~\cite{aldrian2013inverse,romdhani_face_2002}. Often, these steps are separate and applied iteratively. The second class of algorithms traditionally consists of a more complex cost function, using the image information to perform shape from shading, edge detection, and applying a nonlinear solver to jointly or iteratively solve for all the parameters~\cite{romdhani2005estimating,van_rootseler_using_2012,tena_2d_2007,hu2012resolution}. Recently, Sch\"onborn et al. proposed a Markov Chain Monte Carlo based fitting method that integrates automatic landmark detections~\cite{schonborn_monte_2013,egger_pose_2014}. Most of these algorithms require several minutes to fit a single image.

A common point of these algorithms is that they use either only landmark information 
or simple features like raw color values or edge maps, while in a lot of other domains,
most notably 2D facial landmark detection, these have long been superseded by local image 
features like Histogram of Oriented Gradient (HoG)~\cite{dalal2005histograms} or Scale Invariant Feature
Transform (SIFT)~\cite{lowe2004distinctive}. However, it is non-trivial to use 
such features to fit 3D Morphable Models: they are non-differentiable 
operators and have not yet been used for 3DMM fitting.

Recently, cascaded-regression based methods have been widely used with promising results in pose estimation~\cite{dollar2010cascaded,ratsch_wavelet_2012}
and 2D face alignment~\cite{xiong2013supervised,cao2014face,sun2013deep, feng_RCRC_2015}.
In general, a discriminative cascaded regression based method performs optimisation 
in local feature space by applying a learning-based approach to circumvent the problem of non-differentiability. The method allows to learn the gradient of a function from data, instead of differentiating. In this paper, we propose to use a similar strategy to perform 3DMM fitting using SIFT features.

In comparison with existing 3DMM fitting algorithms, using image features and a 
cascaded regression based approach has the potential to give the best of the two worlds:
it is fast, like the linear landmarks-only fitting methods, and at the same time robust
to changing image conditions, and it can be potentially more precise, 
because image information is used to fit shape and pose, and not only landmarks. Furthermore, it is possible to solve for pose and shape parameters simultaneously, 
instead of iteratively. This paper introduces and evaluates the proposed novelty
in the context of fitting pose and shape parameters.

In the rest of this paper, we first give a brief introduction to the cascaded regression method and 3DMMs (Section~\ref{sec:Background}). In Section~\ref{sec:Novelty}, we present the concept of using local image features to fit a 3D Morphable Model, and show how cascaded regression is applied to optimise the Morphable Model parameters in local feature space. We thoroughly evaluate our method using pose and shape data generated from the Morphable Model, as well as on PIE fitting results (Section~\ref{sec:Experiments}). Section~\ref{sec:Conclusion} concludes the paper.

\vspace{-2mm}
\section{Background} \label{sec:Background}
Given an input face image $\mathbf{I}$ and a pre-trained 3DMM, our goal is to find the pose and shape parameters of the 3DMM that best represent the face. In our setting, a face box or facial landmarks are given to perform a rough initialisation of the model. The goal is then to obtain an accurate fitting result using a cost function that incorporates the image information in the form of local features. To facilitate this, the cascaded regression framework is used to learn a series of regressors.

In this section, we will briefly introduce the generic cascaded regression framework and the 3D Morphable Model. The 3D Morphable Model fitting using local image features will then be introduced in Section~\ref{sec:Novelty}.

\subsection{Cascaded Regression} \label{sec:CR}
Given an input image $\mathbf{I}$ and a pre-trained model $\Omega(\boldsymbol\theta)$
with respect to a parameter vector $\boldsymbol\theta$, the aim of a 
regression based method is to iteratively update the parameters
$\boldsymbol\theta \leftarrow \boldsymbol\theta + \delta \boldsymbol\theta$ to
maximise the posterior probability $p{(\boldsymbol\theta|\mathbf{I},\Omega)}$.
A regression based method solves this non-linear optimisation problem 
by learning the gradient from a set of training samples in a supervised manner.
The goal is to find a regressor:
\begin{equation}
\label{single_regressor}
R: \mathbf{f}(\mathbf{I},\boldsymbol\theta) \rightarrow \delta \boldsymbol\theta,
\end{equation}
where $\mathbf{f}(\mathbf{I},\theta)$ is a vector of extracted features from the input image, given the current 
model parameters, and $\delta \boldsymbol\theta$ is the 
predicted model parameter update. This mapping can be learned from a 
training dataset using any regression method, e.g. linear
regression~\cite{xiong2013supervised, feng_RCRC_2015}, 
random forests~\cite{cao2014face} or artificial neural networks~\cite{sun2013deep}.
In contrast to these regression algorithms, a cascaded regression method generates
a strong regressor consisting of $N$ weak regressors in cascade:
\begin{equation}
\label{cascade_regressor}
R = R_1 \circ ... \circ R_N,
\end{equation}
where $R_n$ is the $n$th weak regressor in cascade.
In this paper, we use a simple linear regressor:
\begin{equation}
\label{linear_regression}
R_n: \delta \theta = \mathbf{A}_n\mathbf{f}(\mathbf{I},\theta) + \mathbf{b}_n,
\end{equation}
where $\mathbf{A}_n$ is the projection matrix and $\mathbf{b}_n$ is the 
offset (bias) of the $n$th weak regressor.

More specifically, given a set of training samples $\{\mathbf{f}(\mathbf{I}_i,\theta_i), 
\delta \boldsymbol\theta_i\}_{i=1}^{M}$,
we first apply the ridge regression algorithm to learn the first weak regressor
by minimising the loss:
\begin{equation}
\label{linear_regression_loss}
\sum_{i=1}^M || \mathbf{A}_1 \mathbf{f}(\mathbf{I}_i,\theta_i) + \mathbf{b}_1 - \delta \boldsymbol\theta_i ||^{2} + \lambda ||\mathbf{A}_1||_F^2 ,
\end{equation}
and then update the training samples, i.e. the model parameters and the corresponding feature vectors, using the learned regressor to generate a new training dataset for the second weak regressor learning. This process is repeated until convergence or exceeding a pre-defined maximum number of regressors.

In the test phase, these pre-trained weak regressors are progressively applied to an 
input image with an initial model parameter estimate to update the model and output
the final fitting result. 

\subsection{The 3D Morphable Model}
A 3D Morphable Model consists of a shape and albedo (color) PCA model, of which we use the shape model in this paper. It is constructed from 3D face meshes that are in dense correspondence, that is, vertices with the same index in the mesh correspond to the same semantic point on each face. In 3DMM, a 3D shape is expressed as $\mathbf{v}= [x_1, y_1, z_1, ..., x_{_V}, y_{_V}, z_{_V}]^T$, where $[x_{v}, y_{v}, z_{v}]^T$ are the coordinates of the $v$th vertex and $V$ is the number of mesh vertices. PCA is then applied to this data matrix consisting of $m$ stacked 3D face meshes, which yields $m-1$ eigenvectors $\mathbf{V}_i$, their corresponding variances $\sigma_i^2$, and a mean shape $\mathbf{\bar{v}}$. A face can then be approximated as a linear combination of the basis vectors:
\begin{equation}
\label{eq:pca_face}
\mathbf{v} = \mathbf{\bar{v}} + \sum_{i=1}^{m-1}\alpha_i \mathbf{V}_i,
\end{equation}
where $\boldsymbol{\alpha} = [\alpha_1, ... ,\alpha_{m-1}]^T$ is the shape coefficient or parameter vector.

\section{3D Morphable Model Fitting using Local Image Features} \label{sec:Novelty}
In this section, we will present how we formulate the cascaded regression approach to perform model-fitting using local features.

\subsection{Local Feature based Pose Fitting}

To estimate the pose of the 3D model, we select the parameters vector $\boldsymbol\theta$ to be $ \boldsymbol\theta = [ r_x, r_y, r_z, t_x, t_y, t_z]^T$, with $r_x$, $r_y$, and $r_z$ being the pitch, yaw and roll angle respectively, and $t_x$, $t_y$ and $t_z$ the translations in 3D model space. We can then project a point $\mathbf{p}^{\text{3D}} = [p_x^{\text{3D}}, p_y^{\text{3D}}, p_z^{\text{3D}}, 1]^T$ in homogeneous coordinates from 3D space to 2D using a standard perspective projection:
\begin{equation}
\label{eq:rendering_equation}
\mathbf{p}^{\text{2D}} = \mathbf{P} \times \mathbf{T} \times \mathbf{R}_y \times \mathbf{R}_x \times \mathbf{R}_z \times \mathbf{p}^{\text{3D}},
\end{equation}
where $\mathbf{R}_{x,y,z}$, $\mathbf{T}$ and $\mathbf{P}$ are $4\times 4$ rotation, translation and projection matrices respectively, constructed from the values in $\boldsymbol\theta$, followed by perspective division and converting to screen coordinates.

\begin{figure*}[!t]%
\vspace*{-0.4cm}
\centering
\subfloat[][]{
\includegraphics[width=1.97in,clip=true,trim=0 2 0 55]{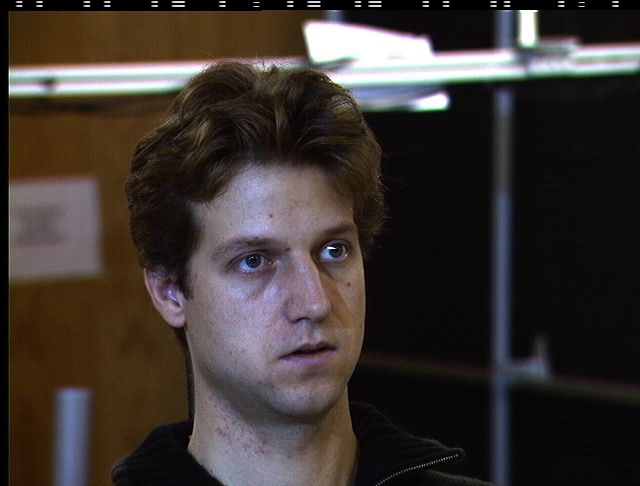}
}%
\qquad
\subfloat[][]{
\includegraphics[width=1.97in,clip=true,trim=0 2 0 55]{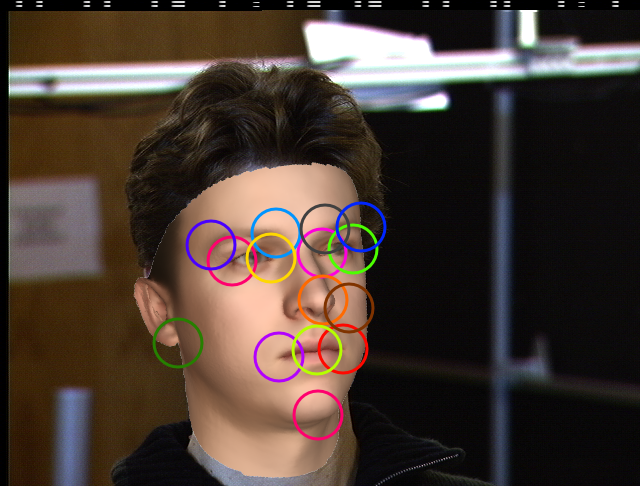} 
}
\qquad
\subfloat[][]{
\includegraphics[width=1.97in,clip=true,trim=0 2 0 55]{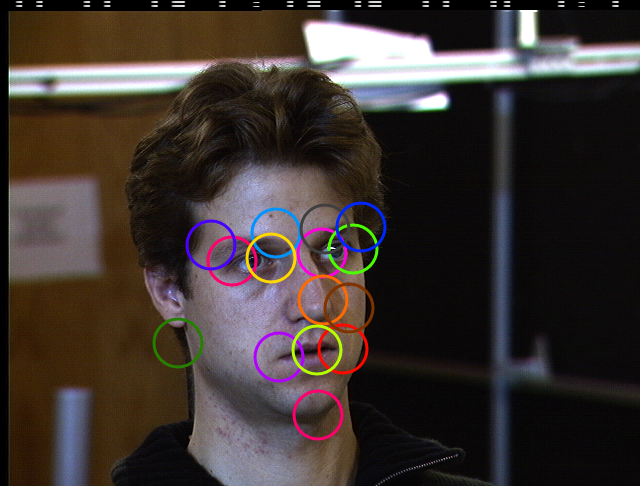}
}
\caption{
3D Morphable Model fitting using local features. 
\textit{(a)} Input image.
\textit{(b)} The 3D Morphable Model is projected using the current set of shape and pose parameters. The circles represent the new locations at which local features are extracted.
\textit{(c)} The local feature regions in the input image, where the features are extracted from. These are used to update the model parameters, and then the process is repeated.
}%
\label{fig:Algorithm}%
\end{figure*}

From the full 3D model's mesh, we choose a subset of $n$ 3D vertices from the mean shape model $\mathbf{\bar{v}}$ in homogeneous coordinates, i.e. $ \mathbf{v}_i \in \mathbb{R}^4 $, $ i \in 0 \dotsc n-1 $. Given the current pose parameters $ \boldsymbol\theta $ we then project them onto the 2D image to obtain a set of 2D coordinates $\mathbf{v}_i^{\text{2D}}$.
Next, local features are extracted from the image around these projected 2D locations, resulting in $n$ feature vectors $\{\mathbf{f}_1, ..., \mathbf{f}_n\}$, where $\mathbf{f}_i \in \mathbb{R}^d$, $d = 128$ in our case of SIFT features. These feature vectors are then concatenated to form one final feature vector, which is the output of $\mathbf{f}(\mathbf{I}, \boldsymbol \theta)$ and the input for the regressor.
Figure~\ref{fig:Algorithm} shows an overview of the process with an input image, the projected 3D model, the locations used to extract local features and their respective location in the input image.

\subsection{Local Feature based Shape Fitting}

As the cascaded regression method allows to estimate arbitrary parameters, we can apply it to estimating the shape parameters in local feature space as well. Our motivation is that the image data contains information about a face's shape, and we want to reconstruct the model's shape parameters for the subject in a given image.
Similar to the previous section, we select a number of 3D vertices $v_i$, but instead of using the mean mesh, we generate a face instance using the current estimated shape coefficients and then use these identity-specific vertex coordinates to project to 2D space.

More specifically, we construct a matrix $ \mathbf{\hat V} \in \mathbb{R}^{3n\times m-1} $ by selecting the rows of the PCA basis matrix $ \mathbf{V} $ corresponding to the $n$ chosen vertex points. A face shape is generated with the formula in equation \ref{eq:pca_face}, using $ \mathbf{\hat V}$ and the current estimate of $\boldsymbol\alpha$. The parameter vector $\boldsymbol\theta$ is then extended to incorporate the shape coefficients: $ \boldsymbol\theta = [ r_x, r_y, r_z, t_x, t_y, t_z, \boldsymbol\alpha]^T $.

Given a new image with a face, and initial landmark locations, initial values for the pose parameter part of $\boldsymbol \theta$ are calculated. The shape initialisation is started at the mean face ($\alpha_i = 0\;\forall i$). The model is projected using this initial estimate, and local features are extracted at the projected 2D locations. Using these features, the regressor predicts a parameter update $\delta \boldsymbol \theta$, and the process is repeated using the new parameter estimate.

While in our case, the points on the mesh we selected coincide with distinctive facial feature points like the eye or mouth corners, any vertex from the 3D model can be used, for example, the points could be spaced equidistant on the 3D face mesh.

It is noteworthy that our method does not rely on particular detected 2D landmarks. A rough initialisation is sufficient to run the cascaded regression. In essence, the proposed method can also be seen as a 3D model based landmark detection, steering the 3D model parameters to converge to the ground truth location. This presents a novel step towards unifying landmark detection and 3D Morphable Model fitting.

\section{Experimental Results} \label{sec:Experiments}
In the following, we present results of the learning based cascaded regression method for pose fitting. Accurate ground truth is 
obtained by generating synthetic data using the 3D Morphable Model.
To simulate more realistic conditions, for each image, a random background is chosen from a backgrounds dataset.

In a second experiment, we perform simultaneous shape and pose fitting using the same method. We subsequently compare the proposed method with POSIT~\cite{dementhon_model-based_1995} and present results on the PIE database.

\subsection{Pose Fitting} \label{sec:ExpPoseFitting}

In this first experiment, we evaluate our method by estimating the 3D pose from the extracted local image features. To train the cascaded regression, we generate poses from $-30^\circ$ to $+30^\circ$ yaw and pitch variation, in $10^\circ$ intervals. Additionally, Gaussian noise with $\sigma = 1.5$mm in x and y translation is added to simulate imprecise initialisation. The model is placed at an initial z-location of $-1200$mm and the focal length was fixed to $1500$. Each sample generated in this way is duplicated and used with 5 different backgrounds. Parameter initialisations are generated by perturbing every angle of each sample with a value uniformly drawn from the interval $[-11^\circ, +11^\circ]$.

\begin{figure}[!t]
\centering
\includegraphics[width=1\linewidth,clip=true,trim=0 8 0 20]{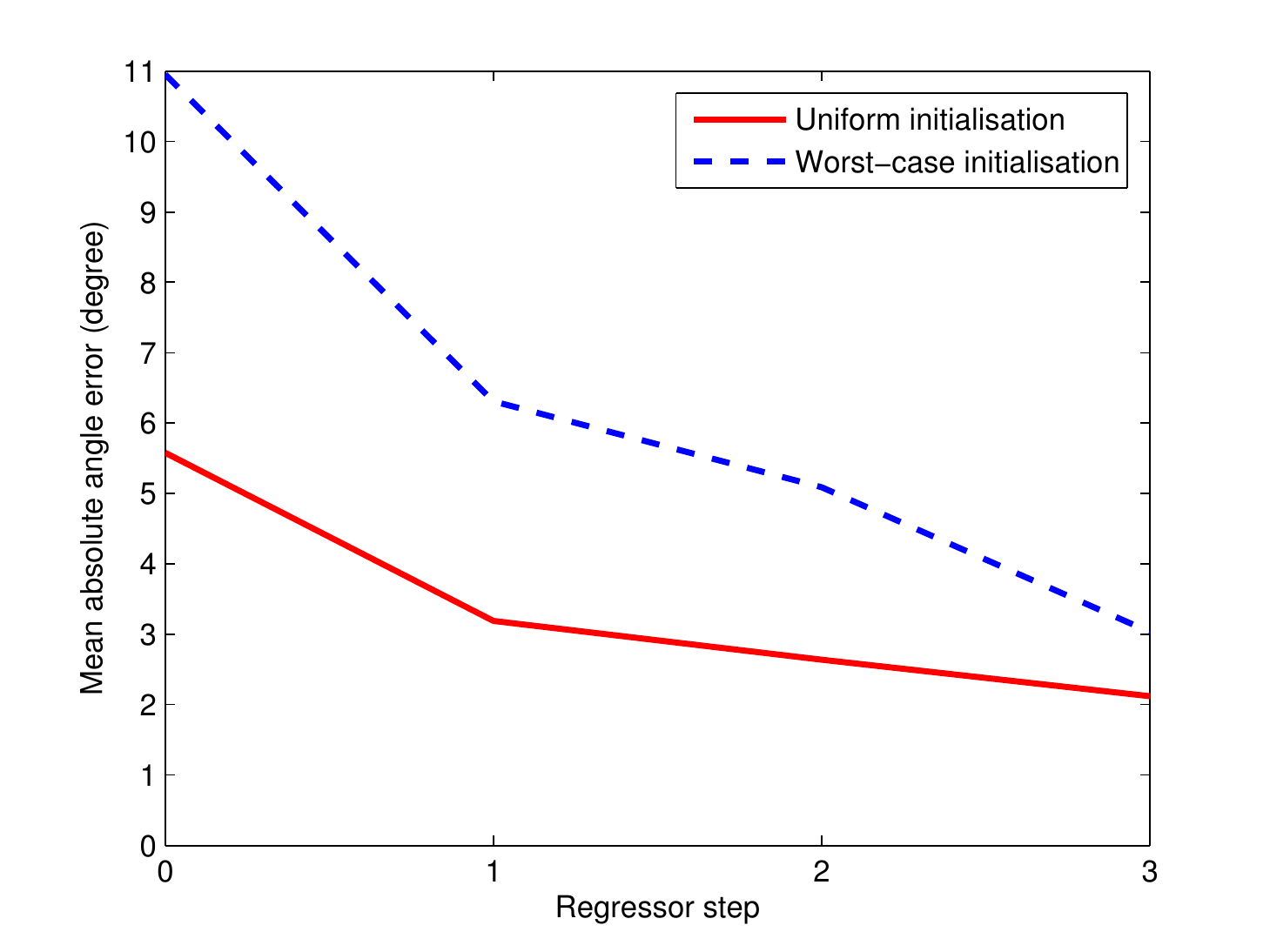} 
\caption{Cascaded regression based 3D Morphable Model fitting. Evaluation of pose fitting under different initialisations. \textit{(red solid):} Initialisation uniformly distributed around the ground truth, \textit{(blue dashed):} Performance in case of bad initialisation.}
\label{fig:SynPose}
\end{figure}

The cascaded regression is tested on the same angle range, but we sample the test data at a finer resolution of $5^\circ$ intervals to verify correct approximation of the learned function.
Figure~\ref{fig:SynPose} shows the mean absolute error of all three predicted angles on the test data. The optimisation is initialised in two different ways: first, with values uniformly drawn in the interval $[-11^\circ, +11^\circ]$ around the ground truth, and second, to evaluate the performance in the case of bad initialisation, the samples are placed $11^\circ$ away from the ground truth in all the images. In both cases, the algorithm converges, and each regressor step promotes further convergence.

\subsection{Simultaneous Shape- and Pose Fitting} \label{sec:ExpShpPose}

The cascaded regression method allows us to simultaneously estimate the shape parameters together with the pose parameters in a unified way. The parameter vector $\theta$ is extended to include the first two PCA shape coefficients of the Morphable Model: $ \boldsymbol\theta = [ r_x, r_y, r_z, t_x, t_y, t_z, \alpha_0, \alpha_1]^T $. We generate test data following the same protocol as described in Section~\ref{sec:ExpPoseFitting}, with the addition that the identity of each test sample is learned as well. 

\begin{figure}[!t]
\centering
\includegraphics[width=1\linewidth,clip=true,trim=0 8 0 20]{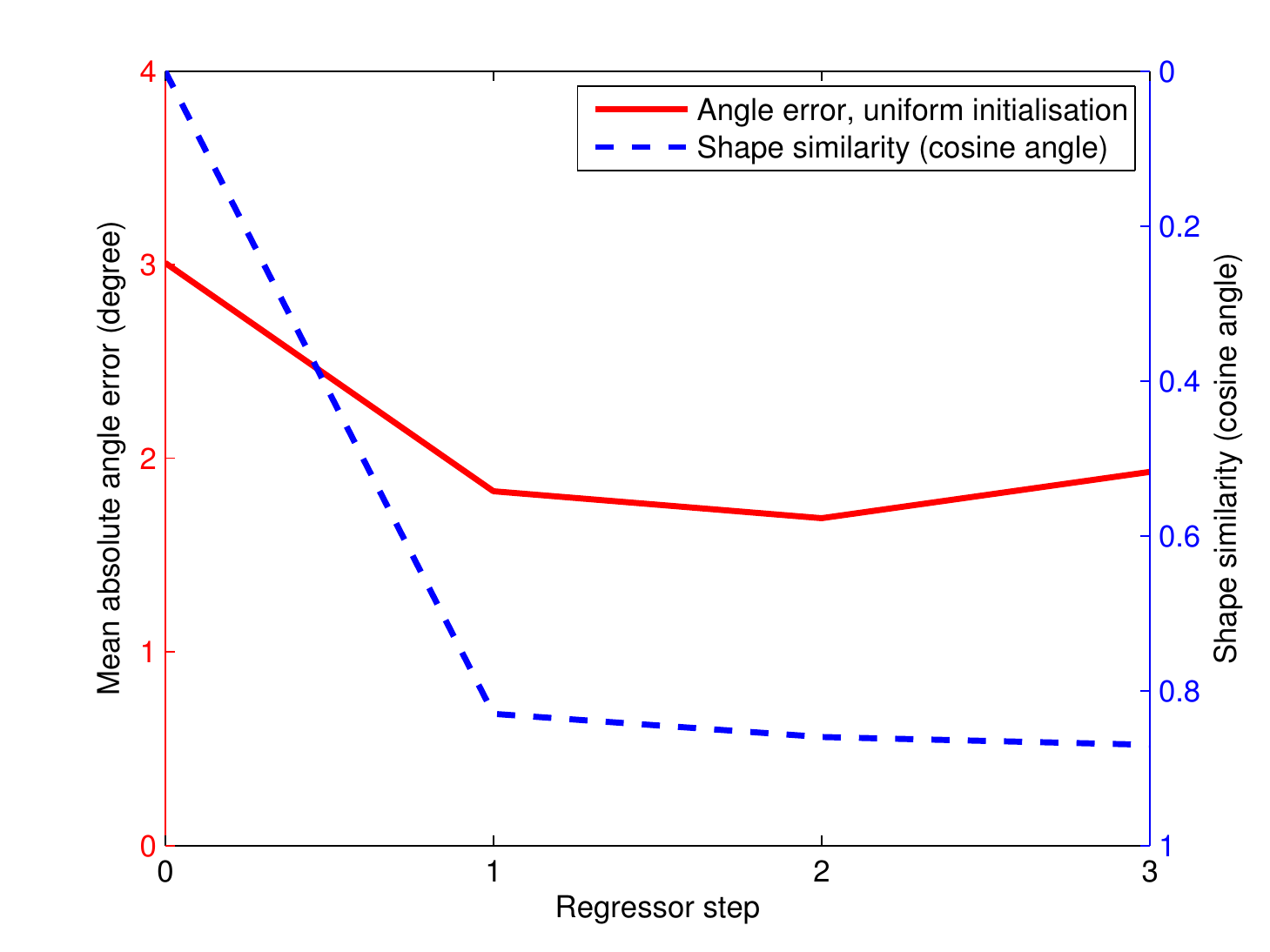} 
\caption{Simultaneous shape- and pose fitting of a 3DMM using cascaded regression. Evaluation on Morphable Model generated data. \textit{(blue dashed):} Shape cosine angle between ground truth and prediction (between 0 and 1, 1 is best), \textit{(red solid):} Mean absolute error of the pose prediction.}
\label{fig:SynShpPose}
\end{figure}

Figure~\ref{fig:SynShpPose} shows the results of the joint shape- and pose fitting. The shape fitting accuracy is measured by the cosine angle between the coefficients of the ground truth $\boldsymbol{\alpha}_g$ and the estimated face $\boldsymbol{\alpha}_e$: $d = \frac{\langle \boldsymbol{\alpha}_e, \boldsymbol{\alpha}_g\rangle}{\|\boldsymbol {\alpha}_e\| \cdot \|\boldsymbol{\alpha}_g\|}$. A shape similarity of 0.87 is achieved, and the pose estimation slightly improves compared to the previous experiment, because the pose can be more accurately estimated when the shape is allowed to change. Similar to the first experiment, this experimental study shows promising results for the local feature based fitting method.

\subsection{Comparison against POSIT}
To relate the performance of our method to existing solutions in the literature, we compare the pose-fitting part of our algorithm against POSIT (Pose from Orthography and Scaling with ITerations). POSIT estimates the pose from a set of 2D to 3D correspondences. In our case, the 2D points are the ground truth landmarks. Tested on the same data as in Section~\ref{sec:ExpPoseFitting}, POSIT achieves a mean absolute error over all angles of $1.84^\circ$, compared to around $2^\circ$ for our method. It should be noted that POSIT achieves these results with very accurate ground truth landmarks, which are often not available in practical applications. If Gaussian noise of 5 pixels is added to the landmarks, the accuracy of POSIT drops to $3.68^\circ$. In contrast to that, the proposed algorithm does not rely on detected landmarks and their accuracy.

\subsection{Evaluation on PIE}

To evaluate the proposed method on real data, we use the fitting results on the PIE database~\cite{sim_cmu_2003} that are provided with the Basel Face Model (BFM,~\cite{bfm09}) as ground truth data. In this way, we can evaluate how well the proposed method can approximate a complex state of the art fitting method.
In particular, to evaluate how the proposed method is able to estimate the shape and pose under changing illumination conditions, we split the available PIE data into illuminations 1 to 17 for training and 18 to 22 for testing, resulting in 3468 training and 1020 test images. Again, a 3-stage cascaded regressor is learned.

\begin{figure}[!t]
\centering
\includegraphics[width=1\linewidth,clip=true,trim=0 8 0 20]{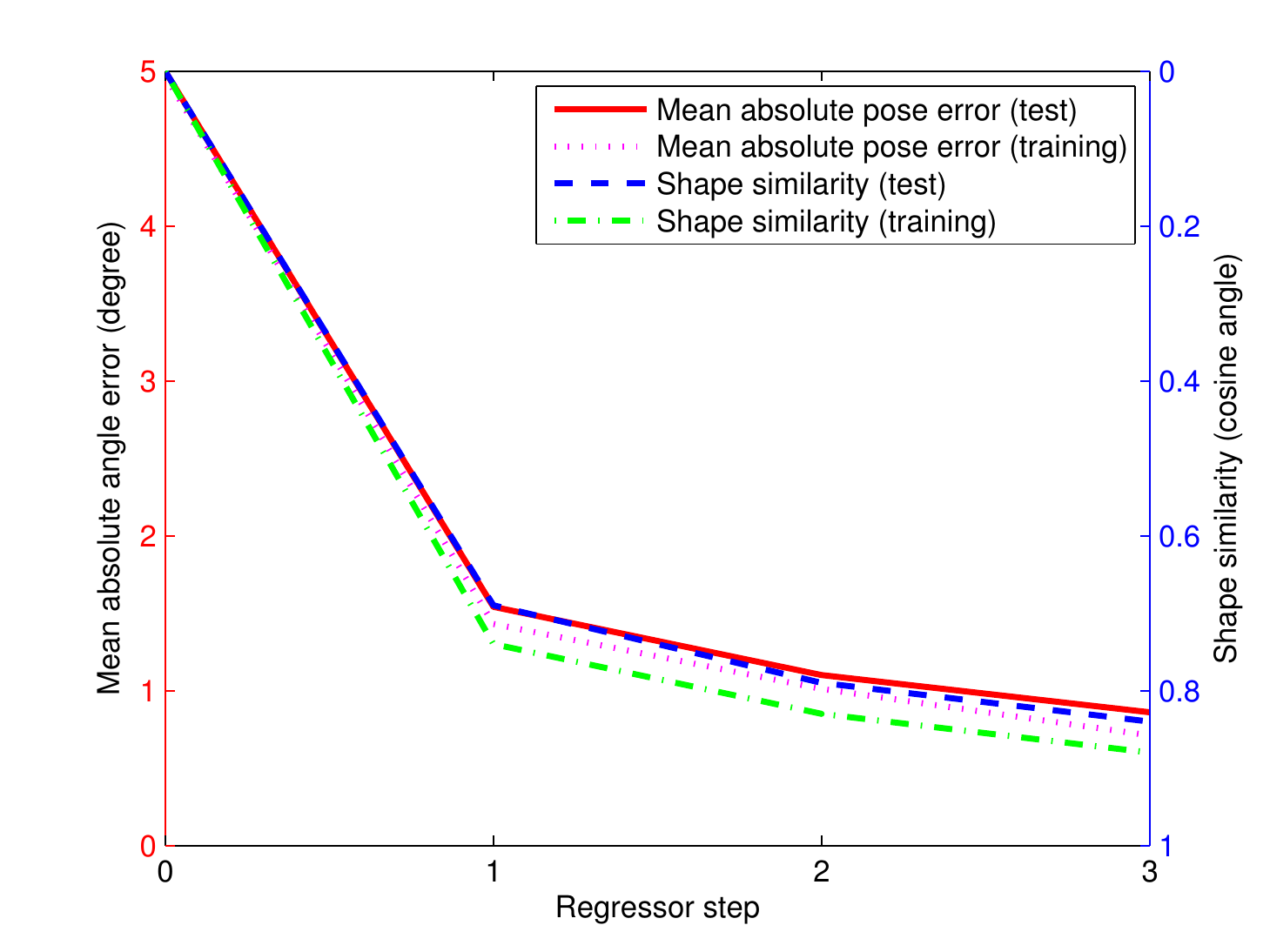} 
\caption{Simultaneous shape- and pose fitting using cascaded regression. Evaluation on the PIE database. \textit{(blue dashed, green dash-dotted):} Shape cosine angle between ground truth and prediction, \textit{(red solid, magenta dotted):} Mean absolute error of the pose prediction over all angles.}
\label{fig:PIE_results}
\end{figure}

Figure~\ref{fig:PIE_results} shows the mean absolute angle error on the training and test set, as well as the cosine angle of the shape coefficients. It can be seen that the test error is only marginally higher than the training error, and the proposed method generalises well to unseen illumination conditions. On the test data, the angle is estimated with an average error of $0.86^\circ$, and a shape similarity of 0.84 is achieved.

\subsection{Runtime}

Cascaded regression methods are inherently fast, and so is the proposed local feature based Morphable Model fitting. Estimating the pose and shape requires around 200 milliseconds per image, with an unoptimised implementation. The runtime largely depends on the number of vertices that are used for local feature extraction, as they directly influence the dimensionality of the feature vector (and thus the prediction time of the regressor). The algorithm could further be sped up by using faster feature extraction or compressing the feature vectors using PCA, but it has, already in its current form, the potential to be used in real-time applications and in a large-scale fashion.

\section{Conclusion} \label{sec:Conclusion}
We proposed a new way to fit 3D Morphable Models that uses local image features. We overcome the obstacle of solving a cost function that contains a non-differentiable feature extraction operator by using a learning-based cascaded regression method that learns the gradient direction from data. We evaluated the proposed method on accurate Morphable Model generated data as well as on real images. The method yields promising results, as local features have been proven robust in many vision algorithms. In contrast to many existing fitting algorithms, our algorithm achieves near real-time performance. Future work includes further evaluation of how the proposed method can unify landmark detection and 3D Morphable Model fitting, and fitting of more shape and possibly albedo coefficients.

\ifCLASSOPTIONcaptionsoff
  \newpage
\fi



\bibliographystyle{IEEEtran}
\bibliography{IEEEabrv,icip.bib}
%



%





\end{document}